\documentclass[11pt,a4paper]{article}
\usepackage[hyperref]{naaclhlt2018}
\usepackage{times}
\usepackage{latexsym}
\usepackage{amsmath,amsthm,amsfonts,amssymb,amsmath,enumitem}
\usepackage{bm}
\usepackage{url}

\usepackage{graphicx}
\usepackage{booktabs}
\usepackage{multirow}
\usepackage{subfig}
\usepackage{tabularx}

\usepackage{footnote}

\aclfinalcopy 
\def\aclpaperid{578} 

\title{Improving Implicit Discourse Relation Classification by Modeling Inter-dependencies of Discourse Units in a Paragraph}

\author{Zeyu Dai, Ruihong Huang \\
	   Department of Computer Science and Engineering \\
       Texas A\&M University \\
  {\tt \{jzdaizeyu, huangrh\}@tamu.edu}}

\date{}

\begin{document}

\maketitle
\begin{abstract}
We argue that semantic meanings of a sentence or clause can not be interpreted independently from the rest of a paragraph, or independently from all discourse relations and the overall paragraph-level discourse structure. 
With the goal of improving implicit discourse relation classification, we introduce a paragraph-level neural networks that model inter-dependencies between discourse units as well as discourse relation continuity and patterns, and predict a sequence of discourse relations in a paragraph. 
Experimental results show that our model outperforms the previous state-of-the-art systems on the benchmark corpus of PDTB.
\end{abstract}

\section{Introduction}
PDTB-style discourse relations, mostly defined between two adjacent text spans (i.e., discourse units, either clauses or sentences), specify how two discourse units are logically connected (e.g., causal, contrast).   
Recognizing discourse relations is one crucial step in discourse analysis and can be beneficial for many downstream NLP applications such as information extraction, machine translation and natural language generation.

Commonly, explicit discourse relations were distinguished from implicit ones, depending on whether a discourse connective (e.g., ``because'' and ``after'') appears between two discourse units \cite{PDTB08}.
While explicit discourse relation detection can be framed as a discourse connective disambiguation problem \cite{pitler2009using,lin2014pdtb} and has achieved reasonable performance (F1 score $>$ 90\%), 
implicit discourse relations have no discourse connective and are especially difficult to identify \cite{Lin2009,lin2014pdtb,xue2015conll}.
To fill the gap, implicit discourse relation prediction has drawn significant research interest recently and progress has been made \cite{chen2016implicit,Liu2016emnlp} by modeling compositional meanings of two discourse units and exploiting word interactions between discourse units using neural tensor networks or attention mechanisms in neural nets. 
However, most of existing approaches ignore wider paragraph-level contexts beyond the two discourse units that are examined for predicting a discourse relation in between.

To further improve implicit discourse relation prediction, we aim to improve discourse unit representations by positioning a discourse unit (DU) in its wider context of a paragraph.
The key observation is that semantic meaning of a DU can not be interpreted independently from the rest of the paragraph that contains it, or independently from 
the overall paragraph-level discourse structure that involve the DU.
Considering the following paragraph with four discourse relations, one relation between each two adjacent DUs:

\noindent(1):\label{P1} {\it [The Butler, Wis., manufacturer went public at \$15.75 a share in August 1987,]$_{DU1}$ \underline{and} \textbf{(Explicit-Expansion)} [Mr. Sim's goal then was a \$29 per-share price by 1992.]$_{DU2}$ \textbf{(Implicit-Expansion)} 
[Strong earnings growth helped achieve that price far ahead of schedule, in August 1988.]$_{DU3}$ \textbf{(Implicit-Comparison)}
[The stock has since softened, trading around \$25 a share last week and closing yesterday at \$23 in national over-the-counter trading.]$_{DU4}$
\underline{But} \textbf{(Explicit-Comparison)} [Mr. Sim has set a fresh target of \$50 a share by the end of reaching that goal.]$_{DU5}$}

Clearly, each DU is an integral part of the paragraph and not independent from other units. 
{\it First}, predicting a discourse relation may require understanding wider paragraph-level contexts beyond two relevant DUs and the overall discourse structure of a paragraph. 
For example, the implicit ``Comparison'' discourse relation between DU3 and DU4 is difficult to identify without the background information (the history of per-share price) introduced in DU1 and DU2.  
{\it Second}, a DU may be involved in multiple discourse relations (e.g., DU4 is connected with both DU3 and DU5 with a ``Comparison'' relation), therefore the pragmatic meaning representation of a DU should reflect all the discourse relations the unit was involved in. 
{\it Third}, implicit discourse relation prediction should benefit from modeling discourse relation continuity and patterns in a paragraph that involve easy-to-identify explicit discourse relations (e.g., ``Implicit-Comparison'' relation is followed by ``Explicit-Comparison'' in the above example).

Following these observations, we construct a neural net model to process a paragraph each time and jointly build meaning representations for all DUs in the paragraph. 
The learned DU representations are used to predict a sequence of discourse relations in the paragraph, including both implicit and explicit relations. 
Although explicit relations are not our focus, predicting an explicit relation will help to reveal the pragmatic roles of its two DUs and reconstruct their representations, which will facilitate predicting neighboring implicit discourse relations that involve one of the DUs.

In addition, we introduce two novel designs to further improve discourse relation classification performance of our paragraph-level neural net model. 
First, previous work has indicated that recognizing explicit and implicit discourse relations requires different strategies, we therefore untie parameters in the discourse relation prediction layer of the neural networks and train two separate classifiers for predicting explicit and implicit discourse relations respectively. 
This unique design has improved both implicit and explicit discourse relation identification performance.  
Second, we add a CRF layer on top of the discourse relation prediction layer to fine-tune a sequence of predicted discourse relations by modeling discourse relation continuity and patterns in a paragraph. 

Experimental results show that the intuitive paragraph-level discourse relation prediction model achieves improved performance on PDTB for both implicit discourse relation classification and explicit discourse relation classification.

\section{Related Work}
\subsection{Implicit Discourse Relation Recognition}
Since the PDTB~\cite{Prasad08thepenn} corpus was created, a surge of studies~\cite{Pitler:2009:ASP:1690219.1690241,Lin2009,Liu2016aaai,rutherford2016robust} have been conducted for predicting discourse relations, primarily focusing on the challenging task of implicit discourse relation classification when no explicit discourse connective phrase was presented. 
Early studies~\cite{Pitler08easilyidentifiable,Lin2009,lin2014pdtb,rutherford2015improving} focused on extracting linguistic and semantic features from two discourse units. 
Recent research~\cite{Zhangemnlp2015,rutherford2016neural,ji2015recursive,ji2016latent} tried to model compositional meanings of two discourse units by exploiting interactions between words in two units with more and more complicated neural network models, including the ones using neural tensor~\cite{chen2016implicit,qin2016stacking,ijcai2017-562} and attention mechanisms~\cite{Liu2016emnlp,lan2017multi,zhou2016attention}. 
Another trend is to alleviate the shortage of annotated data by leveraging related external data, such as explicit discourse relations in PDTB~\cite{Liu2016aaai,lan2017multi,qin2017adversial} and unlabeled data obtained elsewhere~\cite{rutherford2015improving,lan2017multi}, often in a multi-task joint learning framework. 

However, nearly all the previous works assume that a pair of discourse units is independent from its wider paragraph-level contexts and build their discourse relation prediction models based on {\it only} two relevant discourse units. 
In contrast, we model inter-dependencies of discourse units in a paragraph when building discourse unit representations; in addition, we model global continuity and patterns in a sequence of discourse relations, including both implicit and explicit relations.

Hierarchical neural network models~\cite{Liu2017LearningCI,li2016discourse} have been applied to RST-style discourse parsing~\cite{carlson2003building} mainly for the purpose of generating text-level hierarchical discourse structures. In contrast, we use hierarchical neural network models to build context-aware sentence representations in order to improve implicit discourse relation prediction.

\subsection{Paragraph Encoding}
Abstracting latent representations from a long sequence of words, such as a paragraph, is a challenging task. 
While several novel neural network models~\cite{zhang2017deconvolutional,zhang2017spherical} have been introduced in recent years for encoding a paragraph, Recurrent Neural Network (RNN)-based methods remain the most effective approaches.
RNNs, especially the long-short term memory (LSTM)~\cite{hochreiter1997long} models, have been widely used to encode a paragraph for machine translation~\cite{sutskever2014sequence}, dialogue systems~\cite{serban2016building} and text summarization~\cite{nallapati2016abstractive} because of its ability in modeling long-distance dependencies between words.
In addition, among four typical pooling methods (sum, mean, last and max) for calculating sentence representations from RNN-encoded hidden states for individual words, max-pooling along with bidirectional LSTM (Bi-LSTM)~\cite{schuster1997bidirectional} yields the current best universal sentence representation method~\cite{Conneau2017}. 
We adopted a similar neural network architecture for paragraph encoding.

\section{The Neural Network Model for Paragraph-level Discourse Relation Recognition}
\subsection{The Basic Model Architecture}
\begin{figure*}[h]
\includegraphics[height=120mm,width=\textwidth]{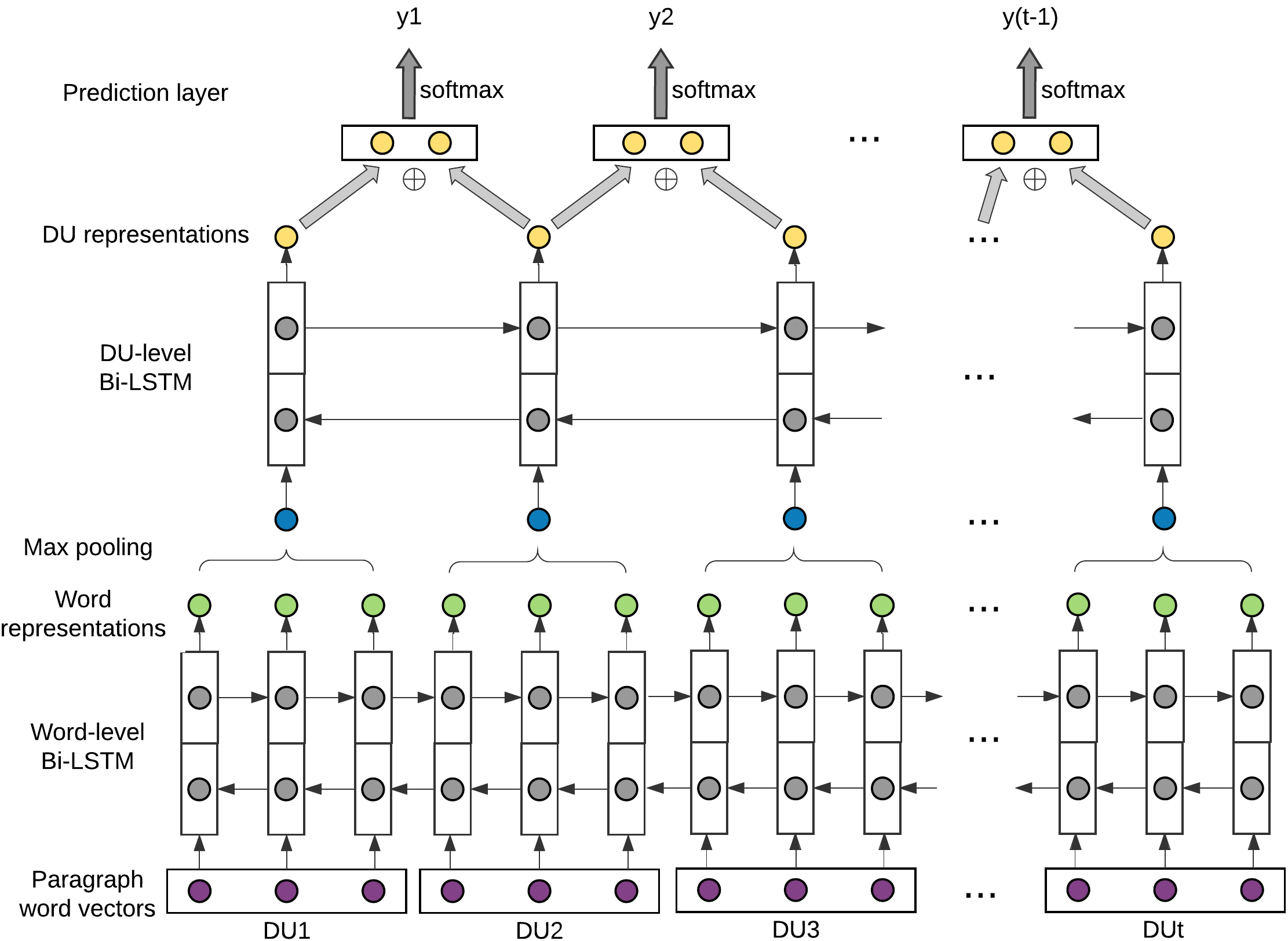}
\caption{The Basic Model Architecture for Paragraph-level Discourse Relations Sequence Prediction.}
\label{paragraph_setting}
\end{figure*}

Figure \ref{paragraph_setting} illustrates the overall architecture of the discourse-level neural network model that consists of two Bi-LSTM layers, one max-pooling layer in between and one softmax prediction layer.
The input of the neural network model is a paragraph containing a sequence of discourse units, while the output is a sequence of discourse relations with one relation between each pair of adjacent discourse units\footnote{In PDTB, most of discourse relations were annotated between two adjacent sentences or two adjacent clauses. For exceptional cases, we applied heuristics to convert them.}. 

Given the words sequence of one paragraph as input, the lower Bi-LSTM layer will read the whole paragraph and calculate hidden states as word representations, and a max-pooling layer will be applied to abstract the representation of each discourse unit based on individual word representations.
Then another Bi-LSTM layer will run over the sequence of discourse unit representations and compute new representations by further modeling semantic dependencies between discourse units within paragraph. 
The final softmax prediction layer will concatenate representations of two adjacent discourse units and predict the discourse relation between them. 

\vspace{.1in}
\noindent{\bf Word Vectors as Input:}
The input of the paragraph-level discourse relation prediction model is a sequence of word vectors, one vector per word in the paragraph.
In this work, we used the pre-trained 300-dimension Google English word2vec embeddings\footnote{Downloaded from \url{ https://docs.google.com/uc?id=0B7XkCwpI5KDYNlNUTTlSS21pQmM}}.
For each word that is not in the vocabulary of Google word2vec, we will randomly initialize a vector with each dimension sampled from the range $[-0.25, 0.25]$.
In addition, recognizing key entities and discourse connective phrases is important for discourse relation recognition, therefore, we concatenate the raw word embeddings with extra linguistic features, specifically one-hot Part-Of-Speech tag embeddings and one-hot named entity tag embeddings\footnote{Our feature-rich word embeddings are of dimension 343, including 300 dimensions for word2vec embeddings  + 36 dimensions for Part-Of-Speech (POS) tags + 7 dimensions for named entity tags. We used the Stanford CoreNLP to generate POS tags and named entity tags.}.

\vspace{.1in}
\noindent{\bf Building Discourse Unit Representations:} 
We aim to build discourse unit (DU) representations that sufficiently leverage cues for discourse relation prediction from paragraph-wide contexts, including the preceding and following discourse units in a paragraph. 
To process long paragraph-wide contexts, we take a bottom-up two-level abstraction approach and progressively generate a compositional representation of each word first (low level) and then generate a compositional representation of each discourse unit (high level), with a max-pooling operation in between. 
At both word-level and DU-level, we choose Bi-LSTM as our basic component for generating compositional representations, mainly considering its capability to capture long-distance dependencies between words (discourse units) and to incorporate influences of context words (discourse units) in each side. 

Given a variable-length words sequence $X = (x_1,x_2,...,x_L)$ in a paragraph, the word-level Bi-LSTM 
will process the input sequence by using two separate LSTMs, one process the word sequence from the left to right while the other follows the reversed direction. 
Therefore, at each word position $t$, we obtain two hidden states $\overrightarrow{h_t}, \overleftarrow{h_t}$. We concatenate them to get the word representation $h_t = [\overrightarrow{h_t}, \overleftarrow{h_t}]$. 
Then we apply max-pooling over the sequence of word representations for words in a discourse unit in order to get the discourse unit embedding:
\begin{center}
\begin{align}
MP_{DU}[j] = \max_{i=DU\_start}^{DU\_end}h_i[j]\quad \\
where, 1 \leq j \leq hidden\_node\_size 
\end{align}
\end{center}

Next, the DU-level Bi-LSTM will process the sequence of discourse unit embeddings in a paragraph and generate two hidden states $\overrightarrow{hDU_t}$ and $\overleftarrow{hDU_t}$ at each discourse unit position. We concatenate them to get the discourse unit representation $hDU_t = [\overrightarrow{hDU_t}, \overleftarrow{hDU_t}]$. 

\vspace{.1in}
\noindent{\bf The Softmax Prediction Layer:}
Finally, we concatenate two adjacent discourse unit representations $hDU_{t-1}$ and $hDU_t$ and predict the discourse relation between them using a softmax function:
\vspace{-.2in}
\begin{center}
\begin{align}
y_{t-1} = softmax(W_y*[hDU_{t-1},hDU_t]+b_y) 
\end{align}
\end{center}
\vspace{-.2in}
\subsection{Untie Parameters in the Softmax Prediction Layer (Implicit vs. Explicit)}
\begin{figure}[h]
\includegraphics[height=42mm,width=80mm]{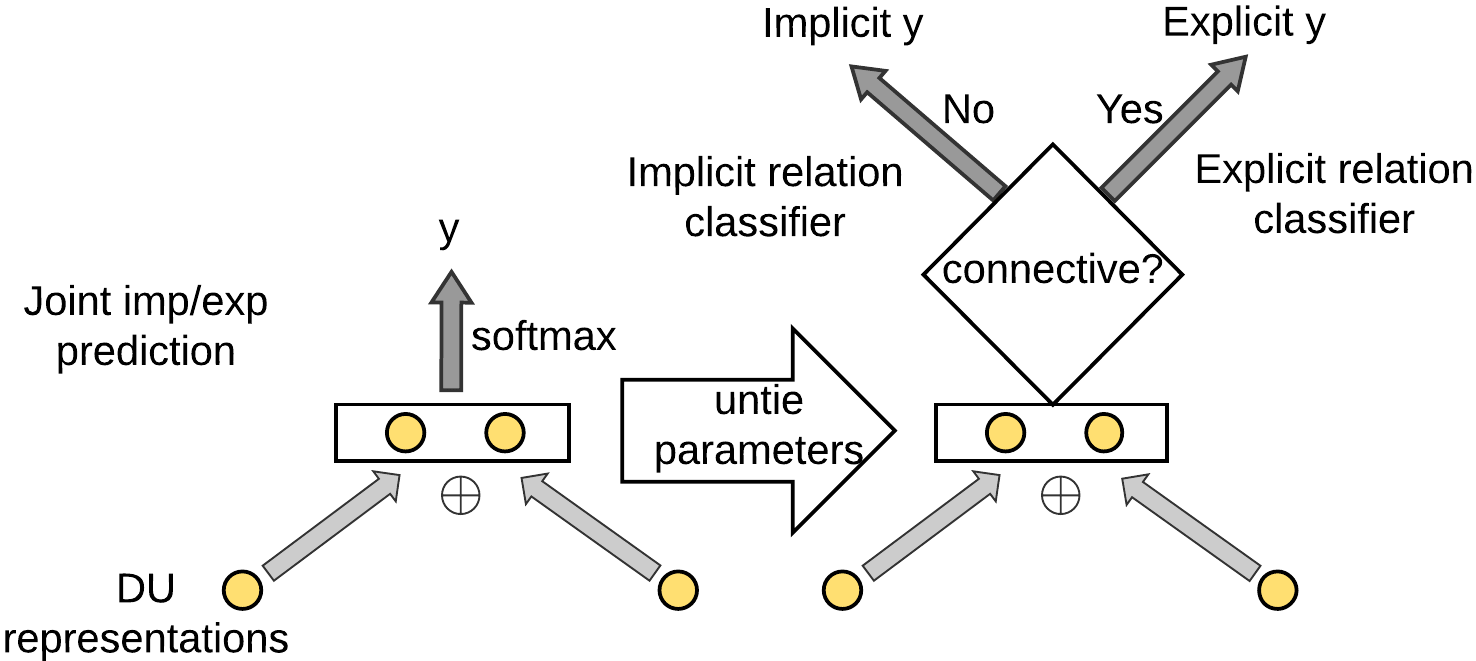}
\caption{Untie Parameters in the Prediction Layer}
\label{untie_output_layer}
\end{figure}

Previous work \cite{pitler2009using,lin2014pdtb,rutherford2016robust} has revealed that recognizing explicit vs. implicit discourse relations requires different strategies. 
Note that in the PDTB dataset, explicit discourse relations were distinguished from implicit ones, depending on whether a discourse connective exists between two discourse units. 
Therefore, explicit discourse relation detection can be simplified as a discourse connective phrase disambiguation problem~\cite{pitler2009using,lin2014pdtb}.
On the contrary, predicting an implicit discourse relation should rely on understanding the overall contents of its two discourse units \cite{lin2014pdtb,rutherford2016robust}.

Considering the different natures of explicit vs. implicit discourse relation prediction, we decide to untie parameters at the final discourse relation prediction layer and train two softmax classifiers, as illustrated in Figure \ref{untie_output_layer}. 
The two classifiers have different sets of parameters, with one classifier for {\it only} implicit discourse relations and the other for {\it only} explicit discourse relations.
\vspace{-.1in}
\begin{center}
\begin{small}
\begin{align}
y_{t-1} = 
\begin{cases}
softmax(W_{exp}[hDU_{t-1},hDU_t]+b_{exp}),&exp\\
softmax(W_{imp}[hDU_{t-1},hDU_t]+b_{imp}),&imp
\end{cases}
\end{align}
\end{small}
\end{center}
\vspace{-.1in}

The loss function used for the neural network training considers loss induced by both implicit relation prediction and explicit relation prediction:
\vspace{-.3in}
\begin{center}
\begin{align}
Loss = Loss_{imp} + \alpha*Loss_{exp} 
\end{align}
\end{center}
The $\alpha$, in the full system, is set to be 1, which means that minimizing the loss in predicting either type of discourse relations is equally important. 
In the evaluation, we will also evaluate a system variant, where we will set $\alpha = 0$, which means that the neural network will not attempt to predict explicit discourse relations and implicit discourse relation prediction will not be influenced by predicting neighboring explicit discourse relations.

\subsection{Fine-tune Discourse Relation Predictions Using a CRF Layer}
Data analysis and many linguistic studies \cite{Pitler08easilyidentifiable,asr2012implicitness,lascarides1993temporal,hobbs1985coherence} have repeatedly shown that discourse relations feature continuity and patterns (e.g., a temporal relation is likely to be followed by another temporal relation). 
Especially, \citet{Pitler08easilyidentifiable} firstly reported that patterns exist between implicit discourse relations and their neighboring explicit discourse relations. 

\begin{figure}[h]
\includegraphics[height=38mm,width=80mm]{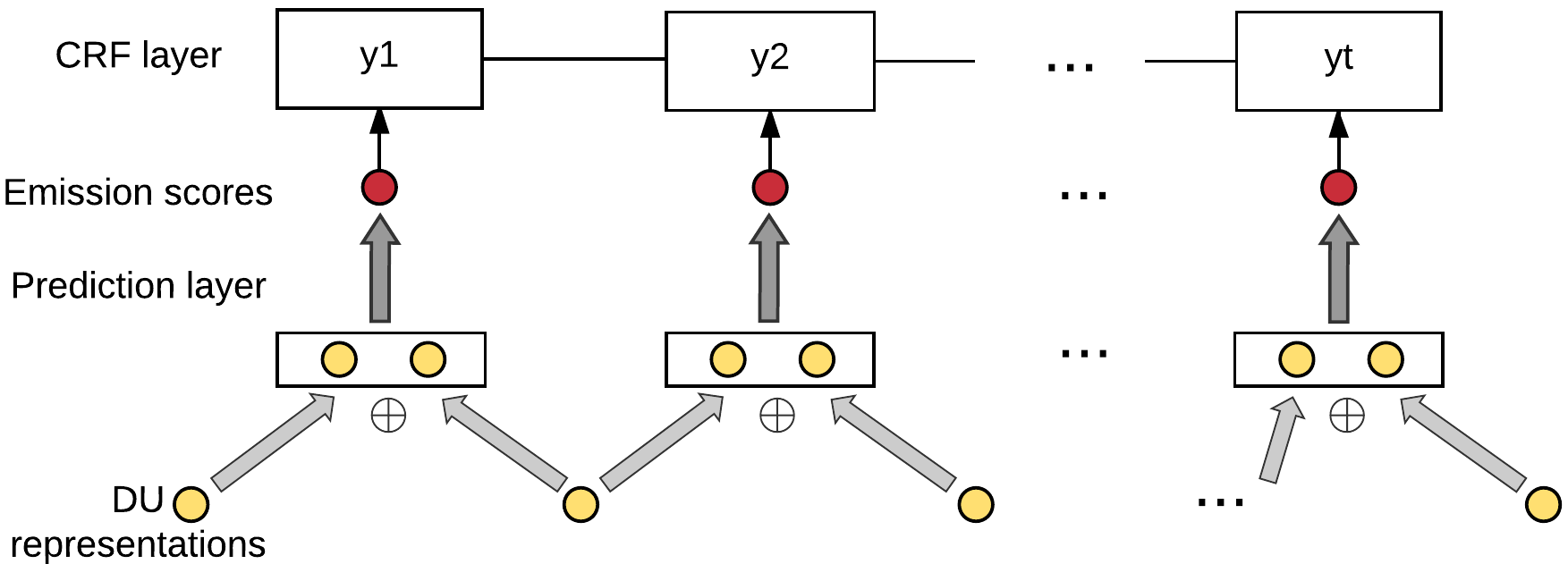}
\caption{Fine-tune Discourse Relations with a CRF layer.}
\label{CRF_layer}
\end{figure}

Motivated by these observations, we aim to improve implicit discourse relation detection by making use of easily identifiable explicit discourse relations and taking into account global patterns of discourse relation distributions. 
Specifically, we add an extra CRF layer at the top of the softmax prediction layer (shown in figure \ref{CRF_layer}) to fine-tune predicted discourse relations by considering their inter-dependencies.

The Conditional Random Fields \cite{lafferty2001conditional} (CRF) layer updates a state transition matrix, which can effectively adjust the current label depending on proceeding and following labels. 
Both training and decoding of the CRF layer can be solved efficiently by using the Viterbi algorithm.
With the CRF layer, the model jointly assigns a sequence of discourse relations between each two adjacent discourse units in a paragraph, including both implicit and explicit relations, by considering relevant discourse unit representations as well as global discourse relation patterns.

\section{Evaluation}
\subsection{Dataset and Preprocessing}
\textbf{The Penn Discourse Treebank (PDTB)}: 
We experimented with PDTB v2.0~\cite{Prasad08thepenn} which is the largest annotated corpus containing 36k discourse relations in 2,159 Wall Street Journal (WSJ) articles. 
In this work, we focus on the top-level\footnote{In PDTB, the sense label of discourse relation was annotated hierarchically with three levels.} discourse relation senses which are consist of four major semantic classes: Comparison (Comp), Contingency (Cont), Expansion (Exp) and Temporal (Temp).
We followed the same PDTB section partition~\cite{rutherford2015improving} as previous work and used sections 2-20 as training set, sections 21-22 as test set, and sections 0-1 as development set. 
Table \ref{PDTB statistics} presents the data distributions we collected from PDTB.

\begin{table}[]
\centering
\begin{tabular}{|c|l|llll|}
\hline
Type                      & Class & Train & Dev & Test & Total \\ \hline
\multirow{4}{*}{Implicit} & Comp     & 1942  & 197 & 152  & 2291  \\
                          & Cont     & 3339  & 292 & 279  & 3910  \\
                          & Exp      & 7003  & 671 & 574  & 8248  \\
                          & Temp     & 760   & 64  & 85   & 909   \\ \hline
\multirow{4}{*}{Explicit} & Comp     & 4184  & 422 & 364  & 4970  \\
                          & Cont     & 2837  & 286 & 213  & 3336  \\
                          & Exp      & 4612  & 481 & 424  & 5517  \\
                          & Temp     & 2742  & 254 & 297  & 3293  \\ \hline
\end{tabular}
\caption{Distributions of Four Top-level Discourse Relations in PDTB.}
\label{PDTB statistics}
\end{table}

\textbf{Preprocessing}: 
The PDTB dataset documents its annotations as a list of discourse relations, with each relation associated with its two discourse units. 
To recover the paragraph context for a discourse relation, we match contents of its two annotated discourse units with all paragraphs in corresponding raw WSJ article. 
When all the matching was completed, each paragraph was split into a sequence of discourse units, with one discourse relation (implicit or explicit) between each two adjacent discourse units\footnote{In several hundred discourse relations, one discourse unit is complex and can be further separated into two elementary discourse units, which can be illustrated as [DU1-DU2]-DU3. We simplify such cases to be a relation between DU2 and DU3.}.
Following this method, we obtained 14,309 paragraphs in total, each contains 3.2 discourse units on average.
Table \ref{arg_count} shows the distribution of paragraphs based on the number of discourse units in a paragraph.

\begin{table}[]
\centering
\begin{tabular}{cccccc}
\hline
\# of DUs & 2    & 3    & 4    & 5     & \textgreater5 \\ \hline
ratio  & 44\% & 25\% & 15\% & 7.3\% & 8.7\%         \\ \hline
\end{tabular}
\caption{Distributions of Paragraphs.}
\label{arg_count}
\end{table}

\begin{table*}[t]
\centering
\begin{tabular}{|l|cc|cccc|cc|}
\hline
 & \multicolumn{6}{c|}{Implicit} & \multicolumn{2}{c|}{Explicit} \\ \hline
\multicolumn{1}{|c|}{Model} & Macro& Acc & Comp & Cont & Exp & Temp & Macro & Acc\\ \hline
~\cite{rutherford2015improving} & 40.50 & 57.10 & - & - & - & - & - & - \\
~\cite{Liu2016aaai} & 44.98 & 57.27 & - & - & - & - & - & - \\
~\cite{Liu2016emnlp} & 46.29 & 57.57 & - & - & - & - & - & - \\
~\cite{ijcai2017-562} & 46.46 & - & - & - & - & - & - & - \\
~\cite{lan2017multi} & 47.80 & 57.39 & - & - & - & - & - & - \\ \hline
\multicolumn{9}{|c|}{DU-pair level Discourse Relation Recognition (Our Own Baselines)} \\ \hline
Bi-LSTM & 40.01 & 53.50 & 30.52 & 42.06 & 65.52 & 21.96 & - & - \\
+ tensors & 45.36 & 57.18 & 36.88 & 44.85 & 68.70 & 30.74 & - & - \\ \hline
\multicolumn{9}{|c|}{Paragraph level Discourse Relation Recognition} \\ \hline
Basic System Variant ($\alpha=0$) & 47.56 & 56.88 & 37.12 & 46.47 & 67.72 & 38.92 & - & - \\
Basic System ($\alpha=1$) & 48.10 & 57.52 & 37.33 & 47.89 & 68.39 & 38.80 & 91.93 & 92.89 \\
+ Untie Parameters & 48.69 & \textbf{58.20} & 37.68 & 49.19 & \textbf{68.86} & 39.04 & \textbf{93.70} & \textbf{94.46} \\
+ the CRF Layer & \textbf{48.82} & 57.44 & \textbf{37.72} & \textbf{49.39} & 67.45 & \textbf{40.70} & 93.21 & 93.98 \\ \hline
\end{tabular}
\caption{Multi-class Classification Results on PDTB. We report accuracy (Acc) and macro-average F1-scores for both explicit and implicit discourse relation predictions. We also report class-wise F1 scores.}
\label{4_way result}
\end{table*}

\subsection{Parameter Settings and Model Training}
We tuned the parameters based on the best performance on the development set.
We fixed the weights of word embeddings during training. 
All the LSTMs in our neural network use the hidden state size of 300. 
To avoid overfitting, we applied dropout \cite{hinton2012improving} with dropout ratio of 0.5 to both input and output of LSTM layers.
To prevent the exploding gradient problem in training LSTMs, we adopt gradient clipping with gradient L2-norm threshold of 5.0.
These parameters remain the same for all our proposed models as well as our own baseline models.

We chose the standard cross-entropy loss function for training our neural network model and adopted Adam \cite{kingma2014adam} optimizer with the initial learning rate of 5e-4 and a mini-batch size of 128\footnote{Counted as the number of discourse relations rather than paragraph instances.}.
If one instance is annotated with two labels (4\% of all instances), we use both of them in loss calculation and regard the prediction as correct if model predicts one of the annotated labels.
All the proposed models were implemented with Pytorch\footnote{\url{http://pytorch.org/}} and converged to the best performance within 20-40 epochs.
 
To alleviate the influence of randomness in neural network model training and obtain stable experimental results, we ran each of the proposed models and our own baseline models ten times and report the average performance of each model instead of the best performance as reported in many previous works.

\subsection{Baseline Models and Systems}
We compare the performance of our neural network model with several recent discourse relation recognition systems that only consider two relevant discourse units. 

\begin{itemize}
\vspace{-.05in}
\item\cite{rutherford2015improving}: improves implicit discourse relation prediction by creating more training instances from the Gigaword corpus utilizing explicitly mentioned discourse connective phrases.
\vspace{-.05in}
\item\cite{chen2016implicit}:  a gated relevance network (GRN) model with tensors to capture semantic interactions between words from two discourse units.
\vspace{-.05in}
\item\cite{Liu2016aaai}: a convolutional neural network model that leverages relations between different styles of discourse relations annotations (PDTB and RST \cite{carlson2003building}) in a multi-task joint learning framework. 
\vspace{-.05in}
\item\cite{Liu2016emnlp}: a multi-level attention-over-attention model to dynamically exploit features from two discourse units for recognizing an implicit discourse relation.
\item\cite{qin2017adversial}: a novel pipelined adversarial framework to enable an adaptive imitation competition between the implicit network and a rival feature discriminator with access to connectives.
\vspace{-.05in}
\item\cite{ijcai2017-562}: a Simple Word Interaction Model (SWIM) with tensors that captures both linear and quadratic relations between words from two discourse units.
\vspace{-.05in}
\item\cite{lan2017multi}: an attention-based LSTM neural network that leverages explicit discourse relations in PDTB and unannotated external data in a multi-task joint learning framework. 
\end{itemize}

\subsection{Evaluation Settings}
On the PDTB corpus, both binary classification and multi-way classification settings are commonly used to evaluate the implicit discourse relation recognition performance.
We noticed that all the recent works report class-wise implicit relation prediction performance in the binary classification setting, while none of them report detailed performance in the multi-way classification setting.
In the binary classification setting, separate ``one-versus-all'' binary classifiers were trained, and each classifier is to identify one class of discourse relations. 
Although separate classifiers are generally more flexible in combating with imbalanced distributions of discourse relation classes and obtain higher class-wise prediction performance, one pair of discourse units may be tagged with all four discourse relations without proper conflict resolution. 
Therefore, the multi-way classification setting is more appropriate and natural in evaluating a practical end-to-end discourse parser, and we mainly evaluate our proposed models using the four-way multi-class classification setting.
 
Since none of the recent previous work reported class-wise implicit relation classification performance in the multi-way classification setting, for better comparisons, we re-implemented the neural tensor network architecture (so-called SWIM in \cite{ijcai2017-562}) which is essentially a Bi-LSTM model with tensors and report its detailed evaluation result in the multi-way classification setting. 
As another baseline, we report the performance of a Bi-LSTM model without tensors as well. 
Both baseline models take two relevant discourse units as the only input.

For additional comparisons, We also report the performance of our proposed models in the binary classification setting.

\begin{table*}[t]
\centering
\begin{tabular}{|c|cccc|}
\hline
Model & Comp & Cont & Exp & Temp \\ \hline
\cite{chen2016implicit} & 40.17 & 54.76 & - & 31.32 \\
\cite{Liu2016aaai} & 37.91 & 55.88 & 69.97 & 37.17 \\
\cite{Liu2016emnlp} & 36.70 & 54.48 & 70.43 & 38.84 \\
\cite{qin2017adversial} & 40.87 & 54.56 & 72.38 & 36.20 \\
\cite{ijcai2017-562} & 40.47 & 55.36 & 69.50 & 35.34 \\
\cite{lan2017multi} & 40.73 & \textbf{58.96} & \textbf{72.47} & 38.50 \\ \hline
\multicolumn{5}{|c|}{Paragraph level Discourse Relation Recognition} \\ \hline
\multicolumn{1}{|l|}{Basic System ($\alpha=1$)} & 42.68 & 55.17 & 68.94 & 41.03 \\
\multicolumn{1}{|l|}{+ Untie Parameters} & \textbf{46.79} & 57.09 & 70.41 & \textbf{45.61} \\ \hline
\end{tabular}
\caption{Binary Classification Results on PDTB. We report F1-scores for implicit discourse relations.}
\label{binary result}
\end{table*}

\begin{table*}[t]
\centering
\begin{tabular}{|l|cc|cc|}
\hline
                                 & \multicolumn{2}{c|}{Implicit} & \multicolumn{2}{c|}{Explicit} \\ \hline
\multicolumn{1}{|c|}{Model}      & Macro         & Acc           & Macro         & Acc           \\ \hline
Basic System ($\alpha=1$)              & 49.92         & 59.08         & 93.05         & 93.83         \\
+ Untie Parameters       & 50.47         & \textbf{59.85}         & 93.95         & 94.74         \\
+ the CRF Layer                            & \textbf{51.84}         & 59.75         & \textbf{94.17}         & \textbf{94.82}         \\ \hline
\end{tabular}
\caption{Multi-class Classification Results of Ensemble Models on PDTB.}
\label{ensemble result}
\end{table*}

\subsection{Experimental Results}
\textbf{Multi-way Classification}:
The first section of table \ref{4_way result} shows macro average F1-scores and accuracies of previous works.
The second section of table \ref{4_way result} shows the multi-class classification results of our implemented baseline systems.
Consistent with results of previous works, neural tensors, when applied to Bi-LSTMs, improved implicit discourse relation prediction performance.
However, the performance on the three small classes (Comp, Cont and Temp) remains low.

The third section of table \ref{4_way result} shows the multi-class classification results of our proposed paragraph-level neural network models that capture inter-dependencies among discourse units.
The first row shows the performance of a variant of our basic model, where we only identify implicit relations and ignore identifying explicit relations by setting the $\alpha$ in equation (5) to be 0.
Compared with the baseline Bi-LSTM model, the only difference is that this model considers paragraph-wide contexts and model inter-dependencies among discourse units when building representation for individual DU.
We can see that this model has greatly improved implicit relation classification performance across all the four relations and improved the macro-average F1-score by over 7 percents.
In addition, compared with the baseline Bi-LSTM model with tensor, this model improved implicit relation classification performance across the three small classes, with clear performance gains of around 2 and 8 percents on contingency and temporal relations respectively, and overall improved the macro-average F1-score by 2.2 percents.

The second row shows the performance of our basic paragraph-level model which predicts both implicit and explicit discourse relations in a paragraph.  
Compared to the variant system (the first row), the basic model further improved the classification performance on the first three implicit relations.
Especially on the contingency relation, the classification performance was improved by another 1.42 percents.
Moreover, the basic model yields good performance for recognizing explicit discourse relations as well, which is comparable with previous best result (92.05\% macro F1-score and 93.09\% accuracy as reported in \cite{Pitler08easilyidentifiable}).

After untying parameters in the softmax prediction layer, implicit discourse relation classification performance was improved across all four relations, meanwhile, the explicit discourse relation classification performance was also improved.
The CRF layer further improved implicit discourse relation recognition performance on the three small classes.
In summary, our full paragraph-level neural network model achieves the best macro-average F1-score of 48.82\% in predicting implicit discourse relations, which outperforms previous neural tensor network models (e.g., \cite{ijcai2017-562}) by more than 2 percents and outperforms the best previous system \cite{lan2017multi} by 1 percent.

\vspace{.05in}
\noindent\textbf{Binary Classification}:
From table \ref{binary result}, we can see that compared against the best previous systems, our paragraph-level model with untied parameters in the prediction layer achieves F1-score improvements of 6 points on Comparison and 7 points on Temporal, which demonstrates that paragraph-wide contexts are important in detecting minority discourse relations. 
Note that the CRF layer of the model is not suitable for binary classification.

\begin{figure*}[h]
\includegraphics[height=63mm,width=0.93\textwidth]{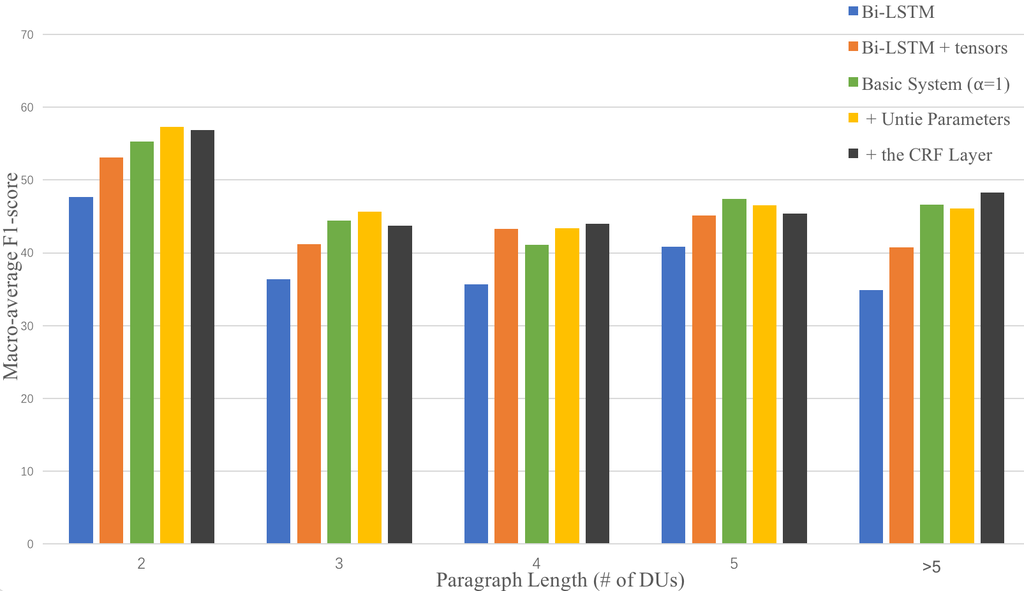}
\caption{Impact of Paragraph Length. We plot the macro-average F1-score of implicit discourse relation classification on instances with different paragraph length.}
\label{impact of paragraph length}
\end{figure*}

\subsection{Ensemble Model}
As we explained in section 4.2, we ran our models for 10 times to obtain stable average performance.
Then we also created ensemble models by applying majority voting to combine results of ten runs.
From table \ref{ensemble result}, each ensemble model obtains performance improvements compared with single model. 
The full model achieves performance boosting of (51.84 - 48.82 = 3.02) and (94.17 - 93.21 = 0.96) in macro F1-scores for predicting implicit and explicit discourse relations respectively. 
Furthermore, the ensemble model achieves the best performance for predicting both implicit and explicit discourse relations simultaneously.

\subsection{Impact of Paragraph Length}
To understand the influence of paragraph lengths to our paragraph-level models, we divide paragraphs in the PDTB test set into several subsets based on the number of DUs in a paragraph, and then evaluate our proposed models on each subset separately.
From Figure \ref{impact of paragraph length}, we can see that our paragraph-level models (the latter three) overall outperform DU-pair baselines across all the subsets. 
As expected, the paragraph-level models achieve clear performance gains on long paragraphs (with more than 5 DUs) by extensively modeling mutual influences of DUs in a paragraph. 
But somewhat surprisingly, the paragraph-level models achieve noticeable performance gains on short paragraphs (with 2 or 3 DUs) as well.
We hypothesize that by learning more appropriate discourse-aware DU representations in long paragraphs, our paragraph-level models reduce bias of using DU representations in predicting discourse relations, which benefits discourse relation prediction in short paragraphs as well.

\subsection{Example Analysis}
For the example (\ref{P1}), the baseline neural tensor model predicted both implicit relations wrongly (``Implicit-Contingency'' between DU2 and DU3; ``Implicit-Expansion'' between DU3 and DU4), while our paragraph-level model predicted all the four discourse relations correctly, which indicates that paragraph-wide contexts play a key role in implicit discourse relation prediction.

For another example:

\noindent(2): {\it [Marshall came clanking in like Marley's ghost dragging those chains of brigades and air wings and links with Arab despots.]$_{DU1}$ \textbf{(Implicit-Temporal)} [He wouldn't leave]$_{DU2}$ \underline{until} \textbf{(Explicit-Temporal)}   
[Mr. Cheney promised to do whatever the Pentagon systems analysts told him.]$_{DU3}$}

Our basic paragraph-level model wrongly predicted the implicit discourse relation between DU1 and DU2 to be ``Implicit-Comparison'',  without being able to effectively use the succeeding ``Explicit-Temporal'' relation.
On the contrary, the full model corrected this mistake by modeling discourse relation patterns with the CRF layer.

\section{Conclusion}
We have presented a paragraph-level neural network model that takes a sequence of discourse units as input, models inter-dependencies between discourse units as well as discourse relation continuity and patterns, and predicts a sequence of discourse relations in a paragraph.
By building wider-context informed discourse unit representations and capturing the overall discourse structure, the paragraph-level neural network model outperforms the best previous models for implicit discourse relation recognition on the PDTB dataset.

\section*{Acknowledgments}
We acknowledge the support of NVIDIA Corporation
for their donation of one GeForce GTX TITAN
X GPU used for this research.

\bibliography{naaclhlt2018}
\bibliographystyle{acl_natbib}


\end{document}